# WORKING MEMORY FOR ONLINE MEMORY BINDING TASKS: A HYBRID MODEL

**Seyed Mohammad Mahdi Heidarpoor Yazdi*a, Abdolhossein Abbassiana**

mahdi.heidarpoor@ipm.ir, abbnet@ipm.ir

* Correspond author

a School of Mathematics, Institute for Research in Fundamental Sciences (IPM), P.O. Box 19395-5746, Tehran, Iran

## ABSTRACT

Working Memory is the brain module that holds and manipulates information online. In this work, we design a hybrid model in which a simple feed-forward network is coupled to a balanced random network via a read-write vector called the interface vector. Three cases and their results are discussed similar to the n-back task called, first-order memory binding task, generalized first-order memory task, and second-order memory binding task. The important result is that our dual-component model of working memory shows good performance with learning restricted to the feed-forward component only. Here we take advantage of the random network property without learning. Finally, a more complex memory binding task called, a cue-based memory binding task, is introduced in which a cue is given as input representing a binding relation that prompts the network to choose the useful chunk of memory. To our knowledge, this is the first time that random networks as a flexible memory is shown to play an important role in online binding tasks. We may interpret our results as a candidate model of working memory in which the feed-forward network learns to interact with the temporary storage random network as an attentional-controlling executive system.

TABLE OF CONTENTS

# 1. INTRODUCTION

The term Working Memory (WM) refers to the brain's module that provides temporary storage and manipulation of information in complex cognition tasks such as learning and abstract reasoning [3,4]. In the language of computer science, WM referred to structures that hold information temporarily for execution [1]; for example, register in computers. However, the "working" part of the WM is not much elaborated. More importantly a good part of this "working" is the binding operation among items which holds in the short-term memory [17]. According to L. Valiant: In "the mind's eye" binding different chunks of memory online is the hallmark of the higher cognitive abilities [23].

Previous works in computational neuroscience have shown the advantages of random networks in input-output processes that require memory [21,26]. Most of these transformations require learning over random networks to decode temporary evolving representations which make them inflexible for WM [6]. Indeed, random networks can inherently maintain information and encoding of input stimulus without any specific learning [6,11]. There is also evidence for similar functioning of WM in the prefrontal cortex [18]. Other models also identify some of these characteristics of working memory for maintaining information but the role of executive functions is less developed [6,16]. In the present work, we take a balanced random network component to be a temporary flexible memory, without any learning, which is separate from the executive function, very much similar to the new hybrid models in machine learning [12,13]. One of the main contributions of the present work is to show the ability of random balanced networks to support binding memory tasks without any learning to take place while passing the stored information to the executive component. This saves much computational work that is usually devoted to storing learned memories in such networks.

An important area of research is the so-called neural binding problem which comprises many distinct problems with different computational and neural requirements. While binding chunks of memory online is regarded as the main objective of WM, a wide range of connectionists approach to such online bindings, were new relations and entities can be dynamically added to a system, fails [9]. On the other hand, the n-back task has become a standard test of executive WM function in attentional control [15]. So, we design simple memory binding tasks, called the first and the second-order memory binding tasks, consisting of a copy of the input and a selected sequence of previous inputs, similar to the n-back task. At the first step, we test the balanced random network as short-term memory in our model then we test the feed-forward network as executive control.

In conformity to our approach, the final example is a cue-based memory binding task in which a cue encodes binding relations and determines which stored chunks of memory should bind to new input. In computer science this mapping a query, our cue, and a key-value pairs to output by executive function is described as attention function [25]. More generally, we also think this goal-driven selection over "the mind's eye" is interpreted in psychology and cognitive sciences as attention [2,19]. This meaningful online selection and decoding of the random network's conjunctive encoding by feed-forward network is our main result regarding the importance of our online binding tasks [19].

The rest of the paper is organized as follows. In the next section, we introduce the components of our hybrid network in more detail. After focusing on the main features of the feed-forward and random balanced networks we also explain how the two components are related through what we define as the interface vector. We then discuss how learning takes place in our network and highlight the main difference to other similar networks. Next in the result section, we explain and analyze the performance of our model on three simple memory binding tasks. In each case, we show the simulation results and discuss the good performance of our hybrid network given what to expect from these simple memory binding tasks. Finally, we show, how the good performance of our proposed hybrid network can be extended to a much more difficult and challenging task such as the cue-based binding task given the nice and powerful capacity of balanced random networks. This will pave the way for addressing more challenging problems such as contextually based working memory models in the near future.

## 2. METHOD

### 2.1. Computational Model

Our hybrid model of Working Memory has two components. A multilayer feed-forward network (FFN) as a controller with learning and a balanced random neural network (BRN) acting as a flexible short-term memory device without learning. (Fig 1)

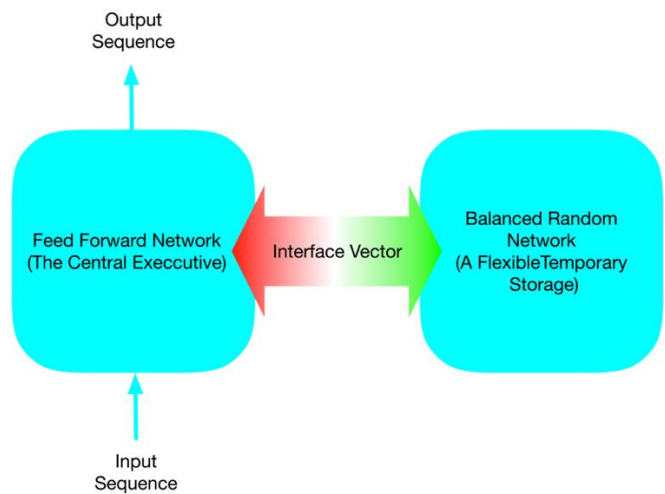

*Fig 1. A hybrid working memory*

### 2.1.1. Feed-Forward Network

The feed-forward network has three hidden layers with a linear activation function and an output layer with a sigmoid activation function.

At each time step a subset of nodes in the output layer, the write vector is shown in green (Fig2), is fed directly to BRN. The links from the third layer to the write vector are not learned and fixed by uniform distribution on [0,1]. All other connections are learned by the backpropagation algorithm. The input layer also consists of two parts, the input vector proper and a read vector, shown in red (Fig1), coming directly from a selected number, c, of BRN nodes. Together the write and the read vectors are represented in the interface vector (IV) between FFN and BRN.

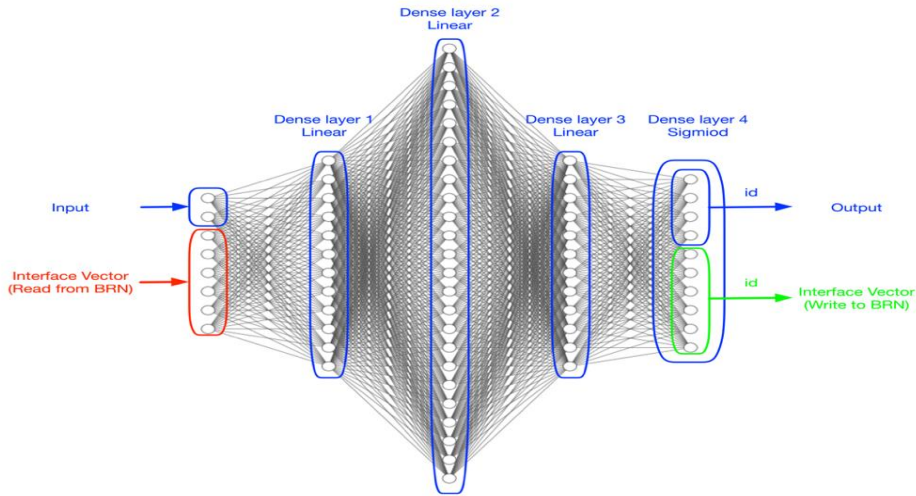

*Fig 2. FFN model*

*A four-layered feed-forward network (FFN) showing the input from BRN (read vector in red) and the output to BRN from the fourth layer (write vector in green).*

The size of FFN is be controlled by parameter controller_dim: it's the number of nodes in layer 1 (the second layer has twice the dim of the first layer and the third layer has the same dim as first). The size of input layer is dim(input)+dim(Interface Vector(IV)) and the size of output layer (fourth layer) is equal to dim(output)+dim(IV). The following condition holds:

$$controller\_dim \geq \dim(IV) + \max\{\dim(input), \dim(output)\}$$

### 2.1.2. Balanced Random Network

At each time step a selected number, c, of nodes receive input from the output layer of FFN (the write vector) while the input layer of the FFN (the read vector) copy the values of the c nodes from the previous time step (the c nodes are fixed throughout the training and test sessions).

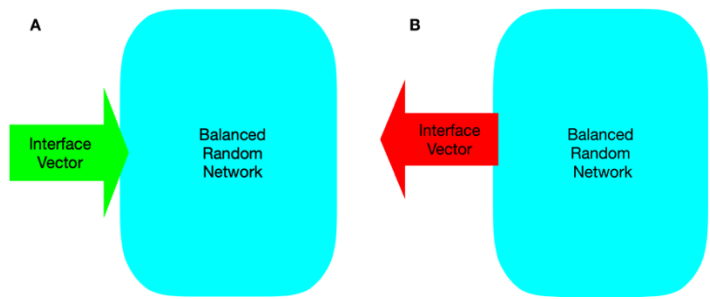

*Fig 3. BRN model*

*Interaction of the balanced random network (BRN) with FFN through the read-write IV. (A) receiving input from FFN (green). (B) sending output to FFN(red).*

We create a random network (RN) with parameters: n=net size, d=average degree of each node, k=Inverse of maximum influence. In other words, RN is a network with size n where each node is randomly connected to d other nodes [8,24]. The weights are such that 2/3 of the connections have value 1/2 (k=2 for positive connection) and the rest of the weights are set to -1 (k=1 for negative connection) [10]. This model of RN has three essential properties: stability, continuity, and orthogonality. For small enough d (d<n/100), RN is also a good transformer [22,27]. The c nodes in RN selected for read and write operation, c<n, are chosen such that $c = \frac{35}{100} n$, similar to the number of direct inputs from sensory to RN in the flexible model of WM [6].

Each node in BRN has dynamics define as follows:

$$P_j^0(t) = \max\left( F * \left( \sum_{i=0}^{n-1} I_j(t) + W_{ji} P_i(t-1) \right), 0 \right)$$

Where $I_j(t)$ is input to node j in BRN (only defined for the c nodes receiving input from FFN), $W_{ji}$ is the connection weight from node i to node j, F is a forget rate parameter which bounds nodes activation and is critical for making RN Balanced (Fig 4).

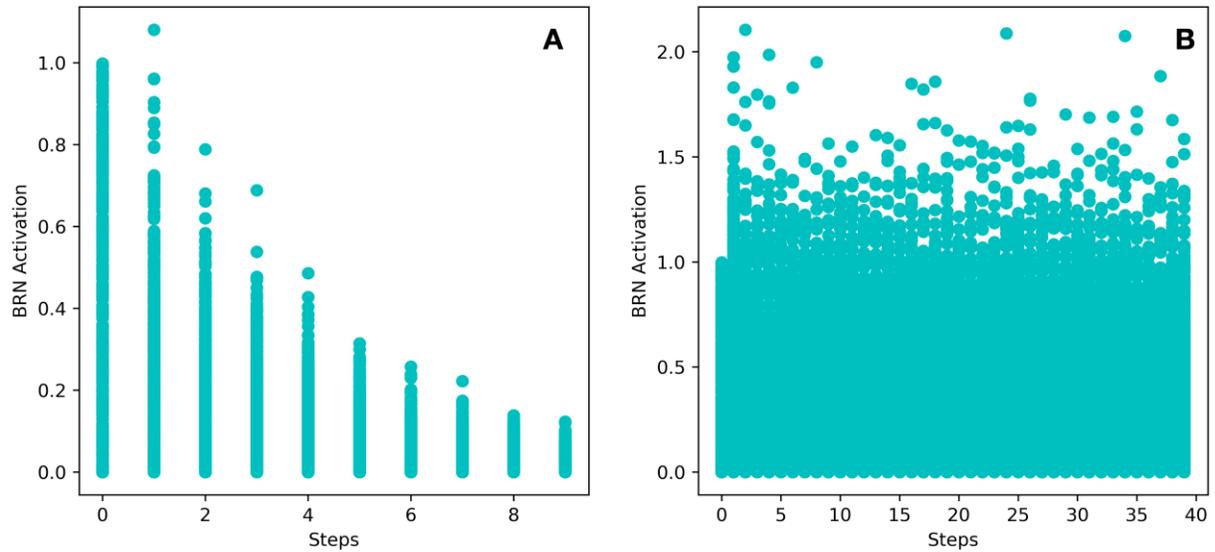

*Fig 4. BRN activation with F=1/3*

*(A) Range of BRN activation after receiving a random firing input at time 0. Activation (information) is maintained for a few steps before reaching the balanced state, low activity. (B) BRN activation under repetitive firing. Activation increases but it's bounded for F=1/3. This helps FFN to extract useful information from BRN.*

### 2.1.3. Interface Vector and the representation of conjunctive encodings

Interface Vector (IV) is an identity map that copies the values of c selected nodes from BRN to FFN's input (Fig 2 and 3. Red) and also copies the value of c number of nodes in the fourth layer of FFN (the non-output nodes) to the same selected c nodes in BRN (Fig 2 and 3. Green).

Here in Fig 5 we see the dynamics of the read-write vector, and show the test results of our "A cue-based memory binding task". Through IV as an identity map, we see how BRN transform input to the output. Although the input shape to BRN is complex, you can see the effect of linear conjunctions in BRN's output. So, BRN encodes inputs to outputs as expected [11,27].

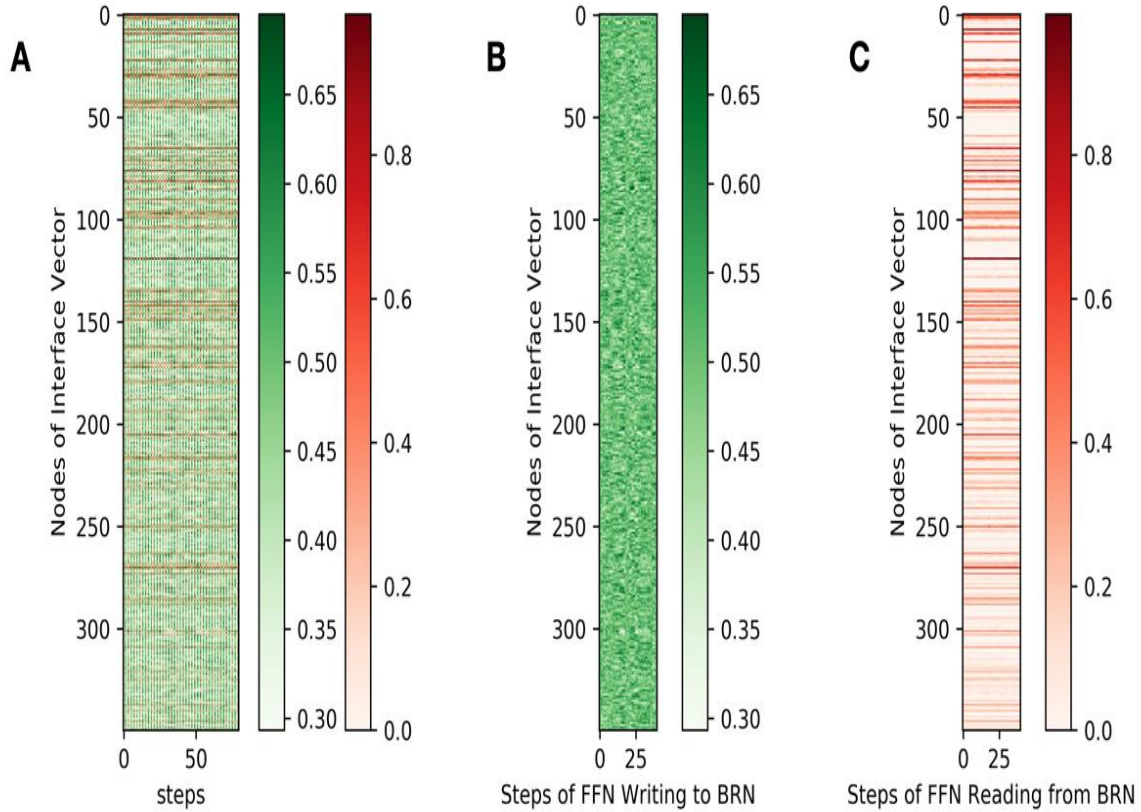

*Fig 5. The read-write after the network is trained for the simple binding task.*

*The read-write in the IV with dimension 350 after the network is trained for the simple binding task. (A) Each column shows the write (green) and read (red) vectors one at a time for a run of more than 50 steps. (B) and (C), showing the same vectors separately.*

## 2.2. Learning

We use a custom model and train our model with a custom train class. We choose RSMprop as the optimizer similar to the optimizer used in similar hybrid models [12,13]. The learning stage has two (or more) phases with different learning rates (LR) 1e-4, 1e-5, and 1e-6 (for sharper results). We tested other optimizers of TF2 such as Adam and obtained similar converging results. RSMprop shows the best convergence speed with our chosen loss function which is mean square error (MSE) in all learning trials.

## 2.3. Testing other Networks

To emphasis some advantages of our model architecture we test a few other networks.

### 2.3.1. Interface Vector and non-trainable connections

Instead of connecting the IV to the fourth layer, we tried a direct connection to layer three to avoid the role of random connections from the third layer to the fourth layer. The results show a much slower convergence (Fig 6). As shown in Fig 7 there is not much interaction with BRN, and indeed, the model cannot learn the task. Comparing the activation levels and the heatmaps it seems does not use BRN in solving the tasks when IV is directly connected to the third layer. Here we emphasize the importance of random connections from the third layer to the fourth layer to generate controlled input to BRN and protect it from sudden stimulus changes that prevent the model from convergence [21].

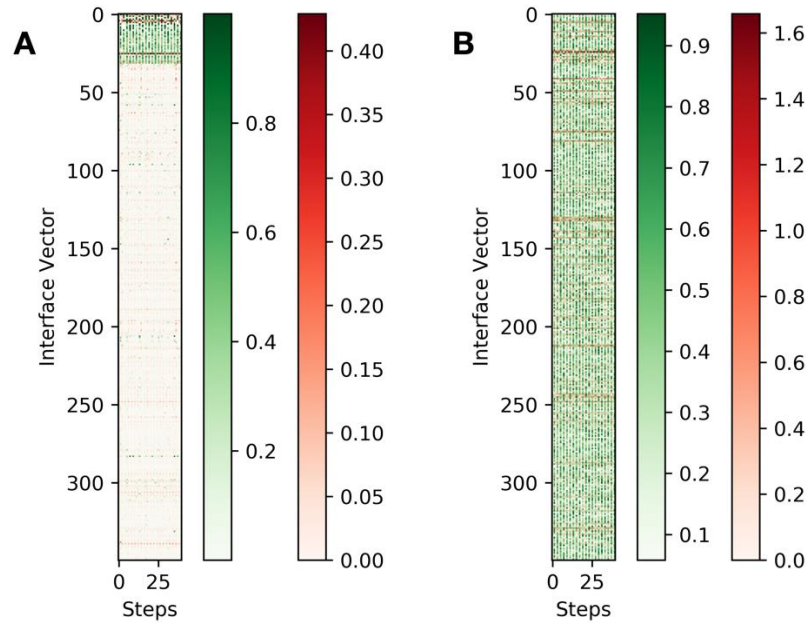

*Fig 7. Comparing read-write with other models*

> *Comparing the read-write activation in test steps after both models are trained in the generalized first-order task (see Fig 6). (A) IV in the other model shows weak interaction between BRN and FFN. (B) For our model, we see a strong interaction between BRN and FFN.*

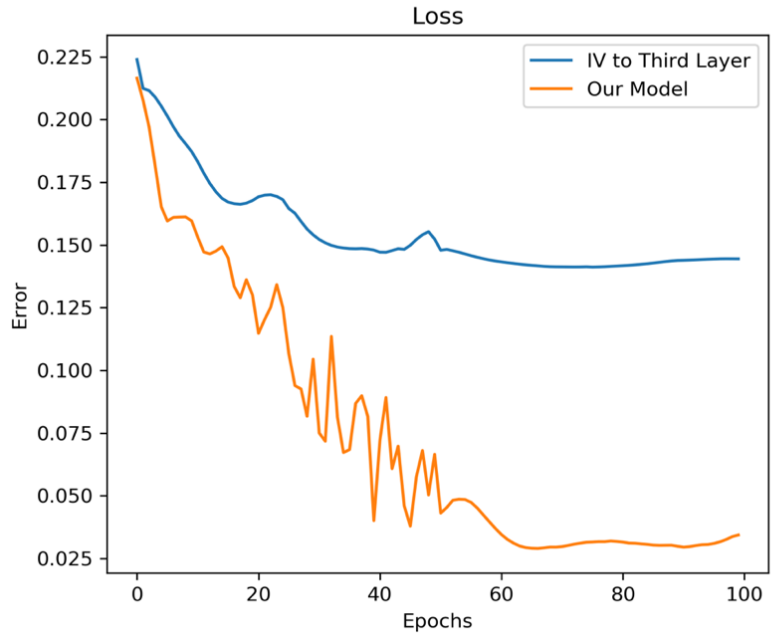

*Fig 6. Train error comparison*

*The convergence is shown for the generalized first-order memory task in our given model (blue curve) versus a different model (orange curve) in which the IV vector is connected directly to the third layer. Here each epoch contains 400 trains, and loss computed by MSE (Mean Square Error). We change LR in epochs=50 from 1e-4 to 1e-5.*

### 2.3.2. Non-random networks

To show the importance of randomness in BRN we test other balanced but non-random networks such as the following: We connect i to j when ($|i - j| < d$) and set the weight to -1 if ($|i - j| \equiv o\ mode\ 3$) otherwise we set the weights to 1/2. The network while is balanced, whole the model cannot learn our tasks. The model shows no convergence and no storage of useful data in the non-random network (Fig 8); This is in contrast to the good property of "linear conjunctive" in RNs that helps to maintain and transform information [6,10,22].

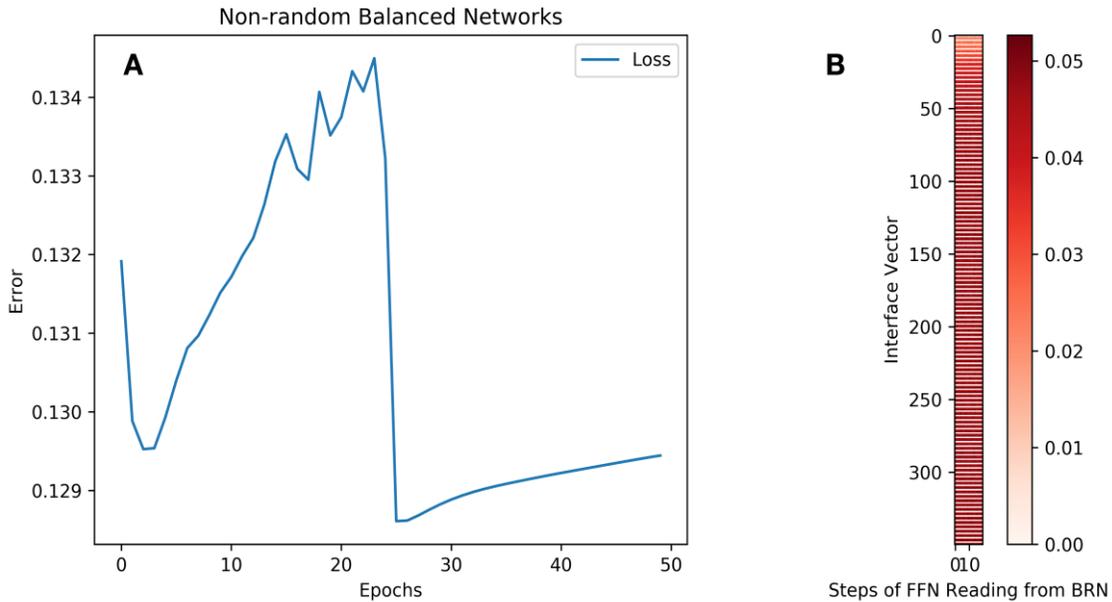

*Fig 8. A balanced but Non-Random network can not learn.*

*A balanced but Non-Random network can not learn the second-order memory task. (A) Loss after each epoch of training (each epoch contains 1000 training samples) does not converge. The sudden fall at step 25 is due to changing the LR but it's big and keeps rising in the following epochs. (B), The IV in reading steps shows low and noninformative activity, as seen in the heat-map, stays near 0(0.05).*

### 2.3.3. RNN and FFN

As is often the case the RNN or the LSTM networks may learn the tasks easily with no biological relevance. They are known to have high convergence rates in learning tasks that are even hard for humans to do. What is important in our case is to have a working memory model similar to humans with minimal components separating the executive (FFN) from the temporary storage component (BRN) which is closed to biologically meaningful representation.

## 2.4. Data and software availability

All simulations were done with python 3.6, TensorFlow 2.1, NumPy, and other well-known libraries. The codes of models are available at Mahdi Heidarpoor GitHub or by contacting MH.

# 3. RESULTS

Our hybrid model of working memory (WM) is a feed-forward network (FFN) as an executive function and a balanced random network (BRN) which acts as a temporary storage unit. The two networks are linked through an interface vector (IV) as shown in Fig1.

We consider three online simple memory binding tasks similar to the n-back task called "first-order memory binding task", "generalized first-order memory task", and "second-order memory binding task". Finally, a more complex memory binding task called "A cue-based memory binding task" is introduced in which a cue is given as input representing a binding relation that prompts the network to choose the useful chunk of memory. Here we deem necessary to make the following remarks. First of all, our model works with time series data which is the keystone of WM [8,23,27]. Second, all tasks require memory but this memory can only be stored in the BRN as the FFN unit has no memory capacity. The network learns to use the information stored in BRN to retrieve the input arrays at the previous time steps.

For all of our tasks, we use a four-layer FFN with controller_dim=512 and BRN with parameters n=1000, d=20, interface_dim=350, and forget_rate=1/3 (see 2 for the exact definitions of these parameters).

## 3.1. First-order memory binding task

In the first-order memory task an input array of $X_1$, $X_2$, $X_3$,… are presented to the FFN at time $t_1$, $t_2$, $t_3$,… respectively. The network learns to connect the present array with the previous one. Each x is a 6-bit array of 0 and 1s and the output a 12-bit array in which the first 6 bits is the exact copy of the input and the second 6 bits is the input presented to the network at the previous time step. For this simple example, there is a total of $2^{12}$ states in which 1000 states are independently chosen at random for training (Fig 9)

As shown in Fig 10, for a sequence 20-time steps given as test inputs, the network has successfully learned to connect the input at time t with the previous input at time t-1. Shifting the blue columns one bit to the left shows an exact match with no error. Note that in all of our tasks the probability of the test sample being in the training set is very low.

## 3.2. Generalized first-order binding task

In the generalized first-order memory task, the model learns to connect input at time t to inputs given at times t-1,…,t-7. Inputs are 4-bit randomly independent generated arrays and the network learns to give a sequence of 32=4*8 bits array of the form ($X_t$, $X_{t-1}$, …, $X_{t-7}$) as the output. In this task, we select 400 independent random data arrays for the training phase and we obtain good convergence results as shown in Fig 11. here, the accuracy of the model for the older data is worse than newer ones (Fig 12).

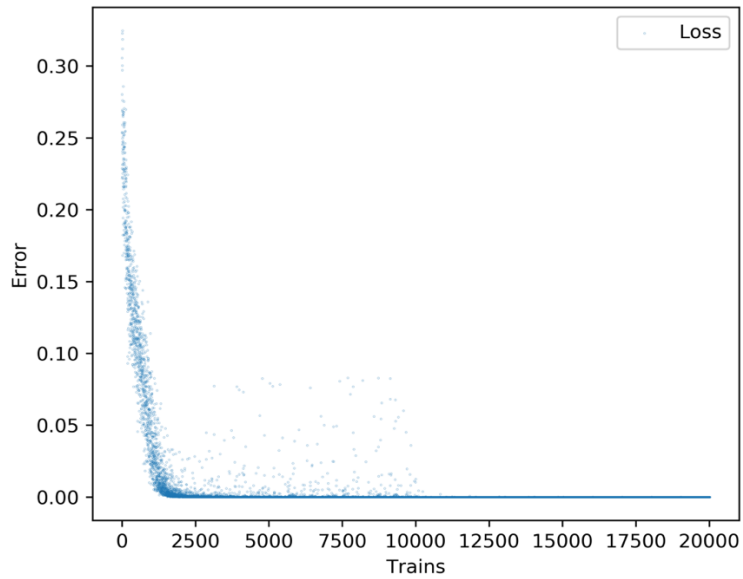

*Fig 9. Train error in the first-order memory task*

*Fast convergence in the first-order memory task. The LR is 1e-4 for the first five epochs(each epoch equals to 1000 training samples) and 1e-5 for all the rest. Error (Loss function) is the Mean Square Error (MSE).*

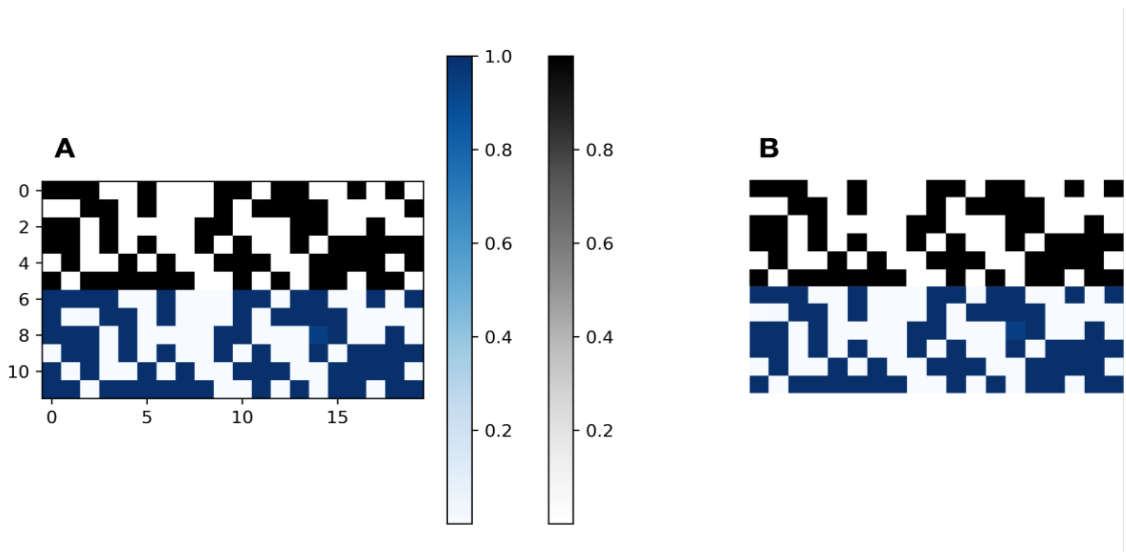

*Fig 10. Outputs for the first-order memory task*

*Outputs (a 12-bit array) for the first-order memory task from 20-time steps. (A) The model learns to bind new input (the 6 bits array shown in black) to the input presented at the previous time step (shown in blue). (B) Shifting the blues to the left by one column makes a perfect match between the black and blue squares.*

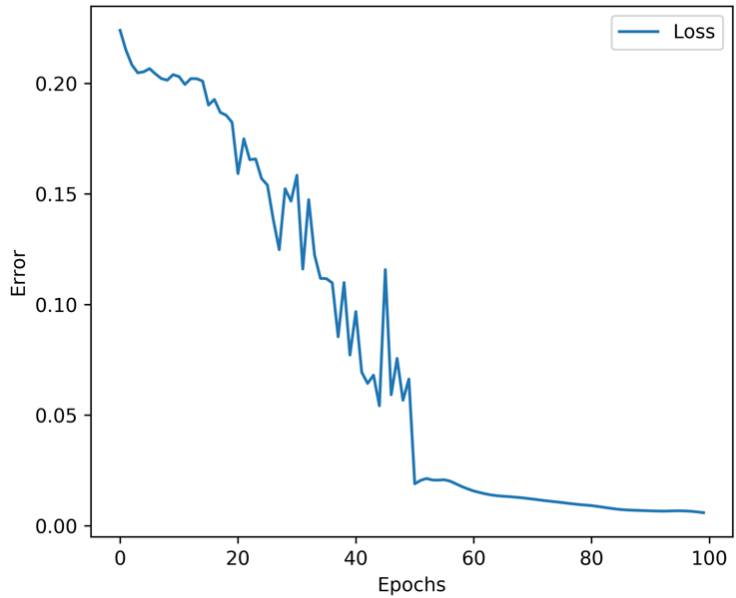

*Fig 11. Train error in the generalized first-order binding task*

*LR changes from 1e-4 to to 1e-5 at epoch 50 (each epoch contains 400 runs). As shown, this change in LR slows down and controls swingings in convergence.*

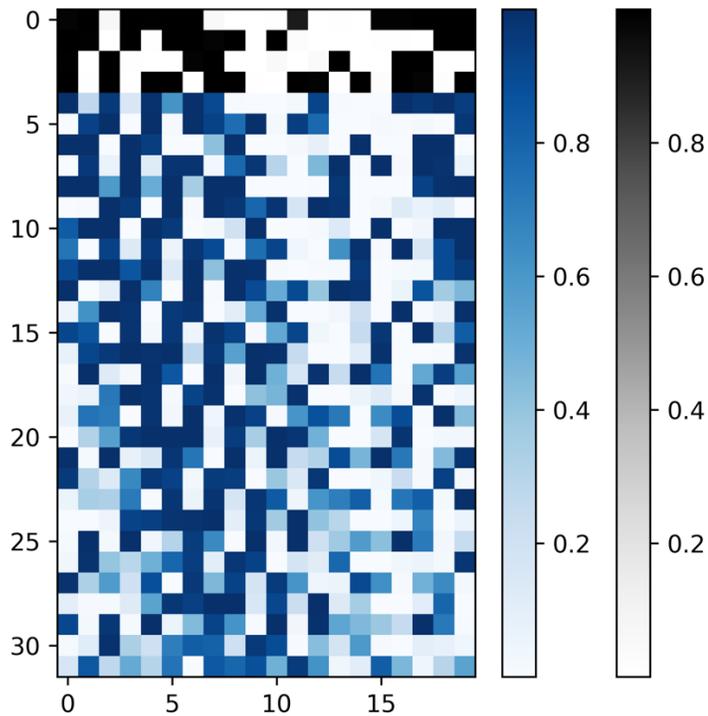

*Fig 12. Outputs of the generalized first-order memory binding task*

*Shown is a 32-bit binary array as the output of the generalized first-order memory task for a period of 20-time steps. Recall of older data gets less accurate as time steps increase.*

## 3.3. Second-order memory binding task

The input at time t is $X_t$, an 8-bit random array, and the output at time t is an array of 16 bits of the form $(X_t, X_{t-2})$. Here the network remembers step t-2 and learns to ignore the input at time step t-1 (Fig 13(A)) online. This can clearly be seen by thresholding the output image in Fig 13(B) and shifting the image columns two bits to the left, Fig 13(C).

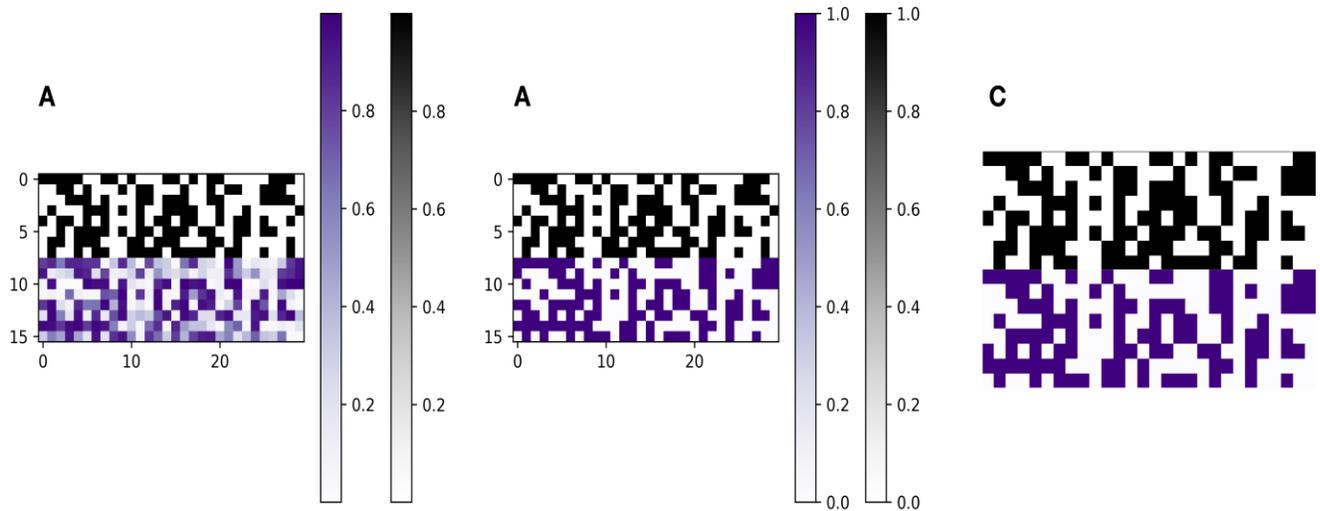

*Fig 13. Outputs for the second-order binding task*

*Output, a 16-bit binary array, for the second-order binding task. (A) The model learns to bind new input (Shown in black) with the input at two previous time steps (purple). (B) Same as (A) after thresholding with threshold=1/2. (C) Shifting the lower purple arrays two columns to the left shows a good match with the upper black arrays.*

Compared to others this is an instance of a more difficult task requiring working memory capabilities involving both retention and manipulation of stored memories. This task is also hard for humans because an extra step of memory storage must take place. We call it second-order memory task to emphasize the importance of this learning ability and the key role played by the BRN component to maintain a memory trace of previous steps.

Here we use 2000 random independent data in the training phase (Fig 14). As you can see the fluctuation in loss function is much larger compared with previous convergence (Fig 9, 11). This could result from BRN sensitivity to inputs and its complex chaotic activity which makes FFN prediction harder to archive [21,26]. Still, this behavior can be controlled by FFN which learns to predict it well. Certainly, this favorable outcome much depends on BRN initialization and non-trainable parameter [10,20,22]. In this respect, we earn different accuracies good enough to learn the tasks (Fig 15).

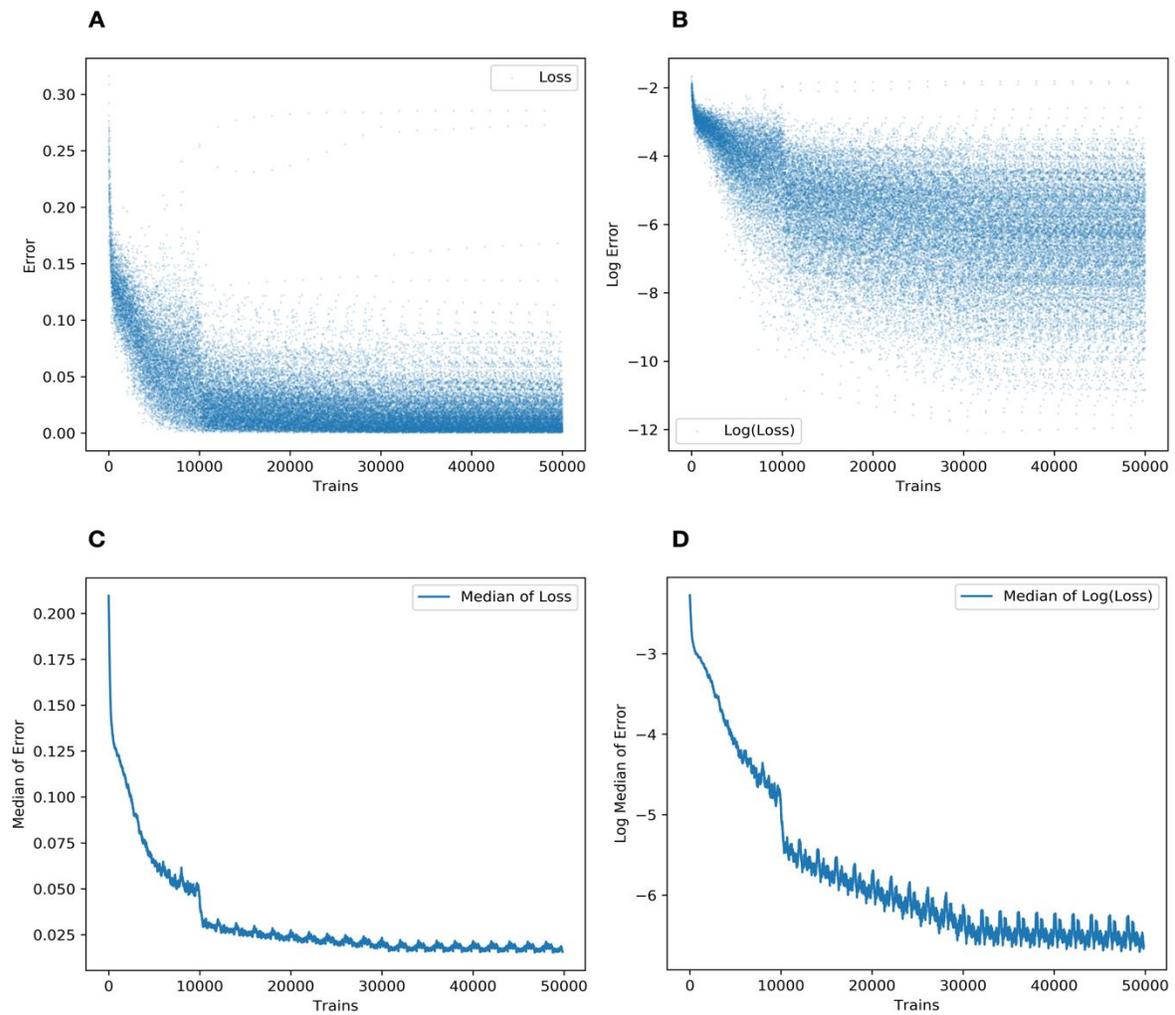

*Fig 14. Train error in the second-order memory task*

*(A) Convergence shows a complex but controllable pattern which results from more dependency on randomness in BRN compared to simple BRN dependency in previous tasks. (B) Shows error in logarithmic scale, where we see the decrease in the magnitude of the error. (C, D) For each training time x the median of x to x+100, which is less than 0.025. The LR Changes from 1e-4 to 1e-5 at epoch 5 (each epoch contains 2000 training samples) and changes from 1e-5 to 1e-6 at epoch 15.*

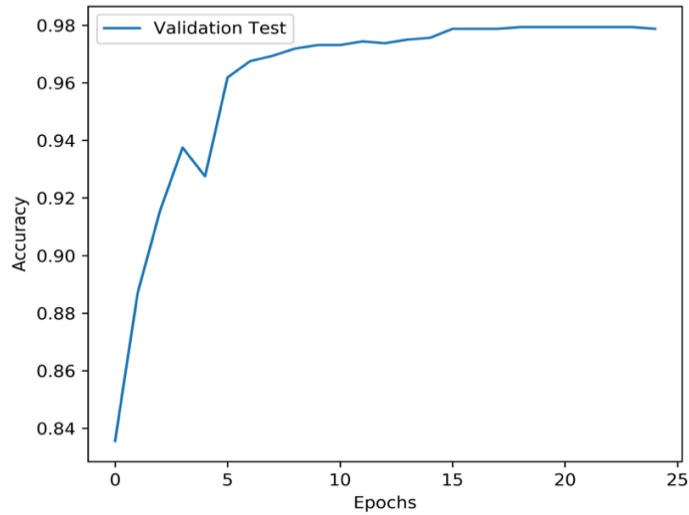

*Figure 15. An Example of a validation test in the second-order memory task*

*We take a validation test on a random set of size 100 after each epoch (=2000 training samples). Accuracy is calculated based on the hamming distance, which is the same as MSE for 0 and 1s.*

### 3.4. A cue-based memory binding task

This is a simple example of a cue-based multi-tasking, nevertheless, the binding involves a more complex operation relative to all the previous tasks. The input is an 8 bits array of the form ($C_t$, $X_t$) where $C_t$ is the online encoding of binding relation of the form 00 or 11 indicating which of the first or the second-order memory binding the model is to perform. Accordingly, the output is of the form ($X_t$, $X_{t-1}$) or ($X_t$, $X_{t-2}$) depending on $C_t$. In other words, the model is to learn which chunks of memory stored in BRN have binding relation with the given input as specified by the given cue. Here, through its learned interaction with BRN, the executive unit (FFN) selects useful chunk of memory stored in BRN online as determined by the given relation. This is similar to what is better known as an instance of a selective attentional mechanism (See 4).

Let's go through this task step-by-step. At time t the input to FFN consists of an 8-bit array, ($C_t$, $X_t$), and the read vector from BRN. For $C_t$=00, we expect the output to be of the form ($X_t$, $X_{t-1}$) and for $C_t$=11, the output is to be of the form ($X_t$, $X_{t-2}$). Inside the model, since no trace of either $X_{t-1}$ or $X_{t-2}$ is stored in FFN this information must be decoded from the read vectors copied from BRN through the IV. Notice that the data carried by IV from the previous step is fixed and independent from the value of $C_t$. So, the FFN must selects and decodes useful information, by learning the binding relation with the given cue. Here, this goal-driven selection which is set by the value of $C_t$ and controlled by the executive function is what we interpret as the top-down selective mechanism (or selective attention) in WM ("the mind's eye") [2,19]. Although being different in many aspects from such mechanisms and not related to our model [1,2]

        To train our model a total of 3000 random and independent set of data arrays were chosen (a somewhat larger data set was used to train the model on the second-order task as it was more difficult to learn compared to the first-order task). As shown in Fig16(A, B) the convergence is much slower with larger swings in the error magnitude compared to the previous tasks. Still, as shown in Fig16(E, F) the results of task validation show good performance when tested on a randomly chosen set of 400 independent input sequences. For the given training data, the model learns the first-order task within the first epoch and with reasonable accuracy (recall error is 22% and goes up a bit up to 23%). Despite more training data, learning the second-order task is much harder and the error stays higher than the first-order task (recall error falls from 43% to 31%). That's the reason we continue the learning trails up to 20 epochs.

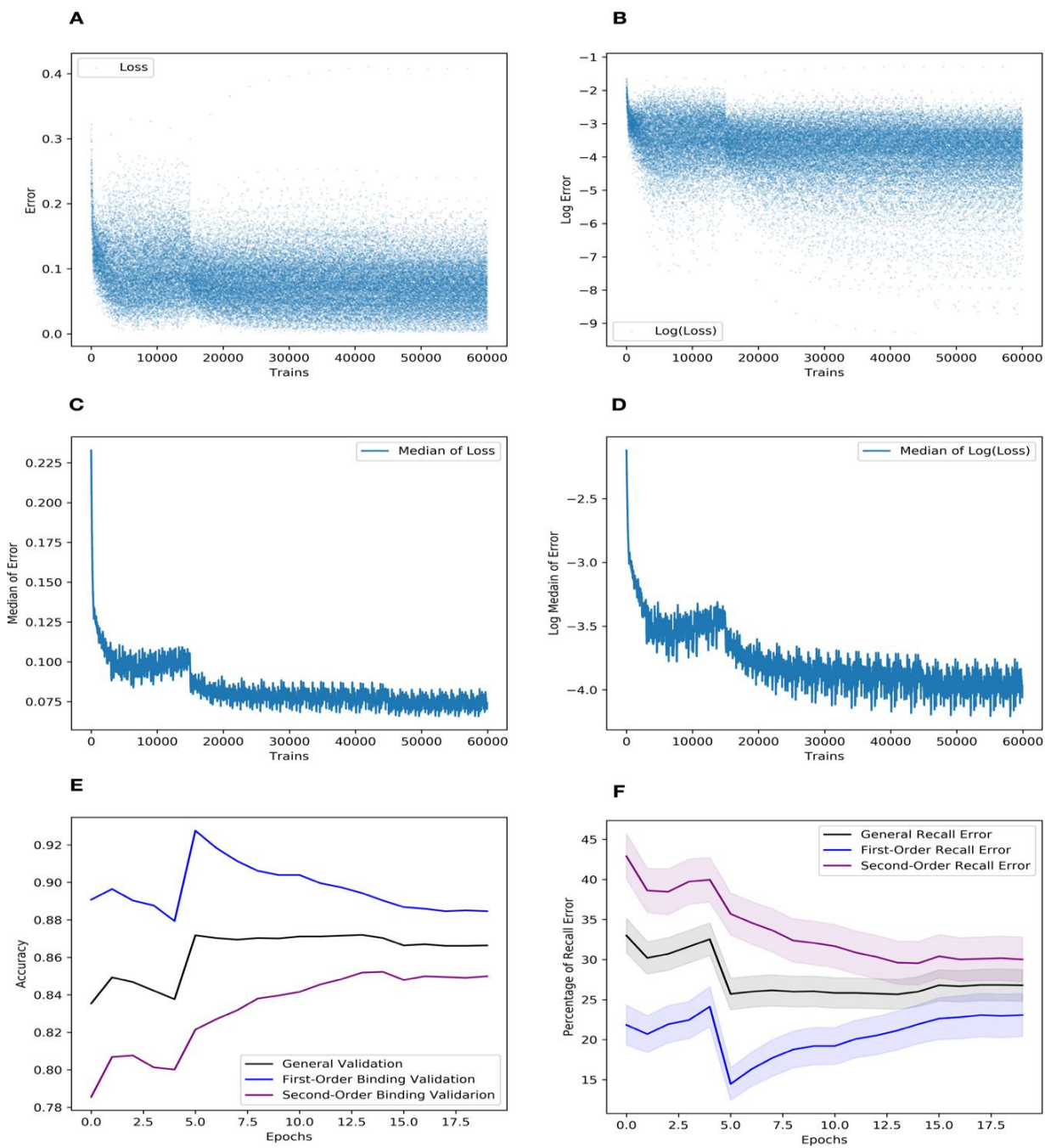

*Fig 16. Error and validation test for the cue-based memory binding task*

*The LR changes from 1e-4 to 1e-5 at epoch 5 and to 1e-5 at epoch 15. The loss diagram shows large swings (A, B) but the median of error from x to x+100 is shown to converges in (C, D). (E) Validation of learned multi-tasking shown for the first-order (blue) and second-order (purple) binding memory tasks based on selective learning cues. (F) General error for the first and the second-order recall with a 95% confidence interval. Notice that the recall error for the more difficult task, second-order memory task, can slowly converge to near 30% while easier first-order memory task convergences very fast and stays below 25% as the number of training data for the second-order task is a bit more than the first-order task.*

Then, Fig17(A, B) shows a test result in which the given cue switches every five steps before and after thresholding. Although not a perfect match we see a large percentage of the output to be correct (green pixels in the bottom half in Fig17(C)). Notice that the set of arrays in the validation phase is created completely independent, random, and is relatively large. A good view of the test phase is illustrated below.

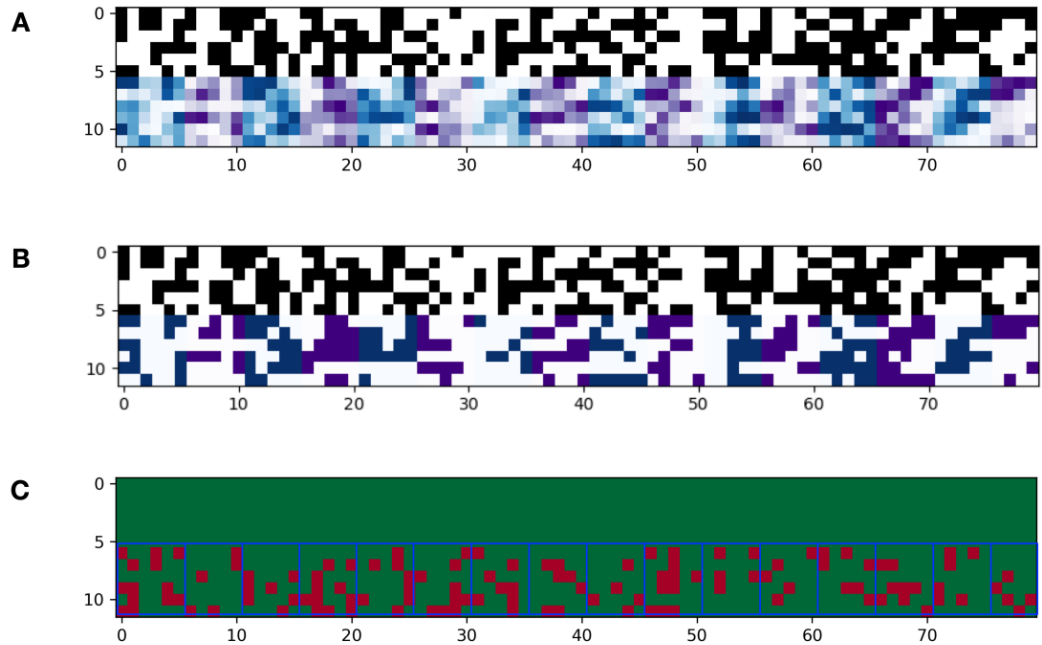

*Fig 17. Outputs of the cue-base memory binding task in test*

*The cue-based memory binding task in test phase. (A) Shows the outputs of the model after the train. Blues show the results of the first-order memory task and Purples show the results of the second-order memory task. (B) Outputs after theresholding. (C) For a better view, we show the difference of model result from the real output of the test (Errors). So, reds show wrong recalls (from memory).*

## 4. DISCUSSION

Artificial neural networks draw much of their inspiration from the biological nervous system. However, the degree of biological relevance is not always the main concern. The good examples are networks such as LSTM or RNN where they sometimes surpass human performance whilst having a superficial resemblance to biological networks. In the new age of artificial intelligence, search to understand NNs' behavior in meaningful ways continues. For example, new hybrid computing models tried to use NNs alongside dynamic external memory. According to Y. Bengio: "These models distinguished keys and values in variations of memory augmented NNs" to emulate reasoning and inference problems in neural language [5,12,13]; but even these suggestive models don't have much biological plausibility [14].

In that regard, our proposed multi-component WM is no exception. The main intention is to separate the roles played by each component in a meaningful way [4]. We create our simple hybrid model to separate temporary memory (BRN) from the executive part (FFN). Notwithstanding, the biological plausibility of BRN and its capability to hold and transform temporary information is validated in many areas of the brain [6,10,11,14,21,22,24,26]. Our basic binding memory tasks aim to show the meaning of these functional features of our model of WM. This is how we think the WM as mental sketchpad can bind information "in mind" online [8,17,23].

The most common requirement of all binding problems is to have a memory storage to hold chunks of the necessary information and online processing of this information [9]. To describe our model better, we design several binding tasks that required temporary memory and online processing. In the case of our first-order memory task, the BRN component is utilized as a simple transformer in which the temporary storage is reduced to its minimum capacity. However, the second-order task tests its capability to maintain information as temporary memory. We also test higher-order memory tasks, third-order and fourth-order, were more difficult to solve and prone to larger errors. The model can solve third-order task. However, when tested on a fourth-order task, binding the present input to the one given at a four-time step before, the model could not solve the task. This is in contrast with our generalized task in that all the system has to do is to bind the present input to the seven previous inputs which are already present in BRN from the last step. This is why we consider the generalized task to be an example of a first-order task. Indeed, the second-order task is much more challenging in that what is given to BRN, say, at time t-1 is to hold the representational content of the arrays ($X_{t-1}$, $X_{t-3}$) whereas at time t the model must connect $X_t$ to $X_{t-2}$. Here, BRN does not simply act as a transformer but must restore $X_{t-2}$ to be recovered at time step t.

In all the previous tasks the FFN component is playing the role of a decoder of the representational content of BRN. In our final more difficult task, we go further and the FFN needs to take one more step and after decoding binding relation that was represented by the given cue (or stimuli), select useful information before decoding the relevant content from BRN. It, therefore, acts as a central executive unit endowed with similar features to common usage of the term attentional function in computer science [19, 25].

## 5. CONCLUSION

Although the binding memory tasks considered in the present study are too abstract and basic, no doubt, we think this is the first time that the very simple hybrid model is capable to solve such online binding tasks. Using the advantage of flexible random networks in meaningful ways that the new entities and relations can be dynamically added which is common in language and abstract reasoning in working memory [4-7,9,13,23]. We believe a deeper analysis of these models' behaviors can give us better insight into some phenomena in neuroscience and psychology, such as attention and selection [6,16,19]. More importantly, most of the new works in computer science have focused on deep learning and needed a lot of computational power and processes [20]. In contrast, we attend to use biological parameters and new structures in computer science, random networks, in order to reduce the computational complexity of our model. Consequently, we hope to extend the capabilities and biological plausibility of our model to solve more challenging binding problems in the future.

## ACKNOWLEDGMENTS

The authors are grateful to Erfan Salavati for helpful discussion. We also thank Massoud Pourmahdian for his useful comments and support throughout this research. This research did not receive any specific grant from funding agencies in the public, commercial, or not-for-profit sectors.

## DECLARATION OF INTERESTS

The authors declare no competing interests.

## REFRENCES